\documentclass[11pt,a4paper]{article}
\usepackage[hyperref]{emnlp-ijcnlp-2019}
\usepackage{times}
\usepackage{latexsym}

\usepackage{url}

\aclfinalcopy

\usepackage{array}
\usepackage{makecell}
\usepackage{amsmath}
\usepackage{pgfplots}

\usepackage{float}

\usepackage{todonotes}
\usepackage{hyperref}
\usepackage{booktabs}

\makeatletter
\newcommand\footnoteref[1]{\protected@xdef\@thefnmark{\ref{#1}}\@footnotemark}
\makeatother

\title{A Richly Annotated Corpus for Different Tasks in Automated Fact-Checking}

\author{Andreas Hanselowski$^{\dagger}$$^*$, Christian Stab$^\dagger$$^*$, Claudia Schulz$^\dagger$$^*$,  \\ 
\textbf{ Zile Li$^*$, Iryna Gurevych$^\dagger$$^*$}  \\[2mm]
   $^\dagger$Research Training Group AIPHES \\
   \url{https://www.aiphes.tu-darmstadt.de} \\[2mm]
   $^*$Ubiquitous Knowledge Processing Lab (UKP-TUDA)\\
   \url{https://www.ukp.tu-darmstadt.de/}\\[1mm]
   $^\dagger$$^*$ Computer Science Department, Technische Universit\"at Darmstadt}

\date{}

\begin{document}
\maketitle

\begin{abstract}
Automated fact-checking based on machine learning is a promising approach 
to identify false information distributed on the web.
In order to achieve satisfactory performance, machine learning methods require a large corpus with reliable
annotations for the different tasks in the fact-checking process.
Having analyzed existing fact-checking corpora, we found that none of them meets these criteria in full.
They are either too small in size, do not provide detailed annotations, or are limited to a single domain.
Motivated by this gap, we present a new substantially sized mixed-domain corpus 
with annotations of good quality for the core fact-checking tasks: 
document retrieval, evidence extraction, stance detection, and claim validation.
To aid future corpus construction, 
we describe our methodology for corpus creation and annotation, and demonstrate that it results in
substantial inter-annotator agreement. 
As baselines for future research, we perform experiments on our corpus with a number of model architectures 
that reach high performance in similar problem settings.
Finally, to support the development of future models, we provide a detailed error analysis for each of the tasks.
Our results show that the realistic, multi-domain setting defined by our data poses new challenges for the existing models, 
providing opportunities for considerable improvement by future systems.
\end{abstract}

%%%%%%%%%%%%%%%%%%%%%%%%%%%%%%%%%%%%%%%%%%%%%%%%%%%%%%%%%%%%%%%%%%%%%%%%%%%%%%%%%%%%%%%%%%%%%%%%%%%
\section{Introduction}

% Motivation of automated methods for fact checking
The ever-increasing role of the Internet as a primary communication channel 
is arguably the single most important development in the media over the past decades. 
While it has led to unprecedented growth in information coverage and distribution speed, it comes at a cost.
False information can be shared through this channel reaching a much wider audience 
than traditional means of disinformation \cite{howell2013digital}.

While human fact-checking still remains the primary method to counter this issue, 
the amount and the speed at which new information is spread makes manual validation challenging and costly. 
This motivates the development of automated fact-checking pipelines \cite{Thorne18Fever,popat2017truth,hanselowski2017framework}
consisting of several consecutive tasks.
The following four tasks are commonly included in the pipeline.
Given a controversial claim, \emph{document retrieval} is applied to identify documents 
that contain important information for the validation of the claim.
\emph{Evidence extraction} aims at retrieving text snippets or sentences from the identified documents 
that are related to the claim. 
This evidence can be further processed via \emph{stance detection} to infer whether it supports or refutes the claim. 
Finally, \emph{claim validation} assesses the validity of the claim given the evidence.

Automated fact-checking has received significant attention in the NLP community in the past years.
Multiple corpora have been created to assist the development of fact-checking models, 
varying in quality, size, domain, and range of annotated phenomena. 
% \Cl{The development of a full-fledged fact-checking system requires several conditions to be met.}
Importantly, the successful development of a full-fledged fact-checking system requires that the underlying corpus satisfies certain characteristics. 
First, training data needs to contain a large number of instances with \textit{high-quality annotations} 
for the different fact-checking sub-tasks.
% \Cl{Since in practice potentially wrong information sources can range from official statements to blog and Twitter posts, 
% the training data should not be limited to a particular domain. }
Second, the training data should not be limited to a particular domain, since potentially wrong information sources can
range from official statements to blog and Twitter posts.

%% Number of claims
We analyzed existing corpora regarding their adherence to the above criteria and identified several drawbacks.
The corpora introduced by \citet{vlachos2014fact, ferreira2016emergent, derczynski2017semeval} 
are valuable for the analysis of the fact-checking problem and provide annotations for stance detection.
However, they contain only several hundreds of validated claims
and it is therefore unlikely that deep learning models 
can generalize to unobserved claims if trained on these datasets.

%% Irrelevant data
A corpus with significantly more validated claims was introduced by \citet{popat2017truth}.
Nevertheless, for each claim, the corpus provides 30 documents 
which are retrieved from the web using the Google search engine instead of a document collection aggregated by fact-checkers. %% 
Thus, many of the documents are unrelated to the claim and important information for the validation may be missing.
% \Revised{This makes it more difficult to train a fact-checking system, as it needs to infer the verdict for a claim
% from a noisy document collection.}{}

The FEVER corpus constructed by \citet{Thorne18Fever} is the largest corpus available 
for the development of automated fact-checking systems. 
It consists of 185,445 validated claims with annotated documents and evidence for each of them.
The corpus therefore allows training deep neural networks for automated fact-checking, 
which reach higher performance than shallow machine learning techniques.
However, the corpus is based 
on synthetic claims derived from Wikipedia sentences 
rather than \emph{natural} claims that originate from heterogeneous web sources. %%
% \Revised{Thus, a system trained on this dataset is unlikely to generalize to a real world fact-checking problem setting.}{}

% our corpus 
In order to address the drawbacks of existing datasets, 
we introduce a new corpus based on the Snopes\footnote{\url{http://www.snopes.com/}} fact-checking website. 
Our corpus consists of 6,422 validated claims with comprehensive annotations 
based on the data collected by Snopes fact-checkers and our crowd-workers. 
The corpus covers multiple domains, including discussion blogs, news, and social media, 
which are often found responsible for the creation and distribution of unreliable information. 
In addition to validated claims, the corpus comprises
over 14k documents annotated with evidence on two granularity levels and with the stance of the evidence with respect to the claims. 
Our data allows training machine learning models for the four steps 
of the automated fact-checking process described above: 
document retrieval, evidence extraction, stance detection, and claim validation.

% our contributions
\noindent The contributions of our work are as follows: 

1) We provide a substantially sized mixed-domain corpus of natural claims with annotations for different fact-checking tasks.
We publish a web crawler that reconstructs our dataset including all annotations\footnote{\url{https://github.com/UKPLab/conll2019-snopes-crawling}}. 
% For a limited circle of people, we are allowed to share the original corpus
For research purposes, we are allowed to share the original corpus\footnote{We crawled and provide the data according to the regulations of the German text and data mining policy. That is, the crawled documents/corpus may be shared upon request with other researchers for non-commercial purposes through the research data archive service of the university library. Please request the data at \url{https://tudatalib.ulb.tu-darmstadt.de/handle/tudatalib/2081}}.

2) To support the creation of further fact-checking corpora, we present our methodology for data collection and annotation, 
which allows for the efficient construction of large-scale corpora with a substantial inter-annotator agreement.

3) For evidence extraction, stance detection, and claim validation we evaluate the performance of high-scoring systems from the
FEVER shared task \cite{thorne2018fact}\footnote{\url{http://fever.ai/task.html/}} and the Fake News Challenge \cite{fnc2017}\footnote{\url{http://www.fakenewschallenge.org/}} as well as the Bidirectional Transformer model BERT~\cite{devlin2018bert} on our data.
To facilitate the development of future fact-checking systems,
we release the code of our experiments\footnote{\url{https://github.com/UKPLab/conll2019-snopes-experiments}}.

4) Finally, we conduct a detailed error analysis of the systems trained and evaluated on our data, 
identifying challenging fact-checking instances which need to be addressed in future research. %by future fact-checking models.

%%%%%%%%%%%%%%%%%%%%%%%%%%%%%%%%%%%%%%%%%%%%%%%%%%%%%%%%%%%%%%%%%%%%%%%%%%%%%%%%%%%%%%%%%%%%%%%%%%%
\section{Related work} \label{sec:relwork}

\begin{table*}%[!b]
\vspace*{-2ex}
\begin{center}
\begin{tabular}{l r r r r r r r }
\toprule
                        & {claims}& {docs.}  & evid. & stance & sources &  rater agr. & domain  \\
\midrule
 PolitiFact14            &    106  &     no  & yes &   no   & no   & no  & political statements  \\
 Emergent16              &    300  &   2,595  &  no &  yes   & yes  & no  & news  \\
 PolitiFact17            &  12,800  &    no   &  no &   no   & no   & no  & political statements   \\
 RumourEval17            &    297  &   4,519  &  no &  yes   & yes  & yes  & Twitter \\
 Snopes17                &   4,956  & 136,085  &  no &   no   & yes  & no  & Google search results \\
 CLEF-2018               &    150  &    no   &  no &   no   & no   & no  & political debates  \\
  FEVER18                & 185,445  &  14,533  & yes &   yes   & yes  & yes  & Wikipedia  \\
  Our corpus             &   6,422  &  14,296  & yes &   yes  & yes  & yes & multi domain  \\
\bottomrule
 \end{tabular}
\end{center}
\vspace{-1.5ex}
\caption{Overview of corpora for automated fact-checking. docs: documents related to the claims; evid.: evidence in form of sentence or text snippets; stance: stance of the evidence; sources: sources of the evidence; rater agr.: whether or not the inter-annotator agreement is reported; domain: the genre of the corpus}	
\vspace{-2.5ex}
    \label{tb:corpora}
\end{table*}

Below, we give a comprehensive overview of existing fact-checking corpora, summarized in Table~\ref{tb:corpora}. 
We focus on their key parameters: fact-checking sub-task coverage, annotation quality, corpus size, and domain.
It must be acknowledged that a fair comparison between the datasets is difficult to accomplish
since the length of evidence and documents, 
as well as the annotation quality, significantly varies between the corpora.

%   Riedel Vlachos, corpus and task definition 
% http://www.aclweb.org/anthology/W14-2508
\noindent\textbf{PolitiFact14} \citet{vlachos2014fact} analyzed the fact-checking problem and constructed a corpus 
on the basis of the fact-checking blog of Channel 4\footnote{\url{http://blogs.channel4.com/factcheck/}} 
and the Truth-O-Meter from PolitiFact\footnote{\url{http://www.politifact.com/truth-o-meter/statements/}}. 
The corpus includes additional evidence,
which has been used by fact-checkers to validate the claims, 
as well as metadata including the speaker ID and the date when the claim was made. 
This is early work in automated fact-checking and \citet{vlachos2014fact} mainly focused on the analysis of the task.
The corpus therefore only contains 106 claims, which is not enough to train high-performing machine learning systems.

% Emergent (FNC) corpus: Even though interesting corpus, not very large
\noindent\textbf{Emergent16} A more comprehensive corpus for automated fact-checking 
was introduced by \citet{ferreira2016emergent}.
The dataset is based on the project Emergent\footnote{\url{http://www.emergent.info/}} 
which is a journalist initiative for rumor debunking.  
It consists of 300 claims that have been validated by journalists.
The corpus provides 2,595 news articles that are related to the claims.
Each article is summarized into a headline and is annotated with the article's stance regarding the claim. 
The corpus is well suited for training stance detection systems in the news domain 
and it was therefore chosen in the Fake News Challenge \cite{fnc2017} for training and evaluation of competing systems.
However, the number of claims in the corpus is relatively small,
thus it is unlikely that sophisticated claim validation systems can be trained using this corpus.

% Liar, Liar Pants on Fire”
% https://arxiv.org/pdf/1705.00648.pdf
\noindent\textbf{PolitiFact17} \citet{wang2017liar} extracted 12,800 validated claims made by public figures in various contexts from Politifact. %\Revised{\footnote{\url{http://www.politifact.com/}}}{}.
For each statement, the corpus provides a verdict and meta information, 
such as the name and party affiliation of the speaker or subject of the debate.
Nevertheless, the corpus does not include evidence 
and thus the models can only be trained on the basis of the claim, the verdict, and meta information. 

% SemEval-2017 Task 8: RumourEval: Determining rumour veracity and support for rumours
\noindent\textbf{RumourEval17} \citet{derczynski2017semeval} organized the RumourEval shared task,
for which they provided a corpus of 297 rumourous threads from Twitter, comprising 4,519 tweets.
The shared task was divided into two parts, \emph{stance detection} and \emph{veracity prediction} of the rumors, 
which is similar to claim validation. 
The large number of stance-annotated tweets allows for training stance detection systems 
reaching a relatively high score of about 0.78 accuracy.
However, since the number of claims (rumours) is relatively small, and the corpus is only based on tweets,
this dataset alone is not suitable to train generally applicable claim validation systems. 

% Where the Truth Lies
\noindent\textbf{Snopes17} A corpus featuring a substantially larger number of validated claims was introduced by \citet{popat2017truth}. 
It contains 4,956 claims annotated with verdicts
which have been extracted from the Snopes website
as well as the Wikipedia collections of proven hoaxes\footnote{\url{https://en.wikipedia.org/wiki/List_of_hoaxes\#Proven_hoaxe}} 
and fictitious people\footnote{\url{https://en.wikipedia.org/wiki/List_of_fictitious_people}}.
For each claim, the authors extracted about 30 associated documents using the Google search engine, 
resulting in a collection of 136,085 documents.
However, since the documents were not annotated by fact-checkers,
irrelevant information is present 
and important information for the claim validation might be missing.

% Shared task on political debates
\noindent\textbf{CLEF-2018} Another corpus concerned with political debates was introduced by \citet{clef2018checkthat:overall}
and used for the CLEF-2018 shared task.
The corpus consists of transcripts of political debates in English and Arabic and provides annotations for two tasks:
identification of check-worthy statements (claims) in the transcripts, and
validation of 150 statements (claims) from the debates.
However, as for the corpus \textbf{PolitiFact17}, no evidence for the validation of these claims is available.

% FEVER corpus
\noindent\textbf{FEVER18} The FEVER corpus introduced by \citet{Thorne18Fever} 
is the largest available fact-checking corpus, consisting of 185,445 validated claims.
The corpus is based on about 50k popular Wikipedia articles.
Annotators modified sentences in these articles to create the claims 
and labeled other sentences in the articles, which support or refute the claim, as evidence. 
The corpus is large enough to train deep learning systems able to retrieve evidence from Wikipedia. 
Nevertheless, since the corpus only covers Wikipedia and the claims are created synthetically,
the trained systems are unlikely to be able to extract evidence from heterogeneous web-sources
and validate claims on the basis of evidence found on the Internet.

% our approach
As our analysis shows, while multiple fact-checking corpora are already available, 
no single existing resource provides full fact-checking sub-task coverage backed by a substantially-sized and validated dataset 
spanning across multiple domains. 
To eliminate this gap, we have created a new corpus as detailed in the following sections.

%%%%%%%%%%%%%%%%%%%%%%%%%%%%%%%%%%%%%%%%%%%%%%%%%%%%%%%%%%%%%%%%%%%%%%%%%%%%%%%%%%%%%%%%%%%%%%%%%%%
\section{Corpus construction} \label{sec:cc}

This section describes the original data from the Snopes platform, 
followed by a detailed report on our corpus annotation methodology.

\subsection{Source data}

\begin{figure}[h]
\fbox{\includegraphics[width=7.5cm]{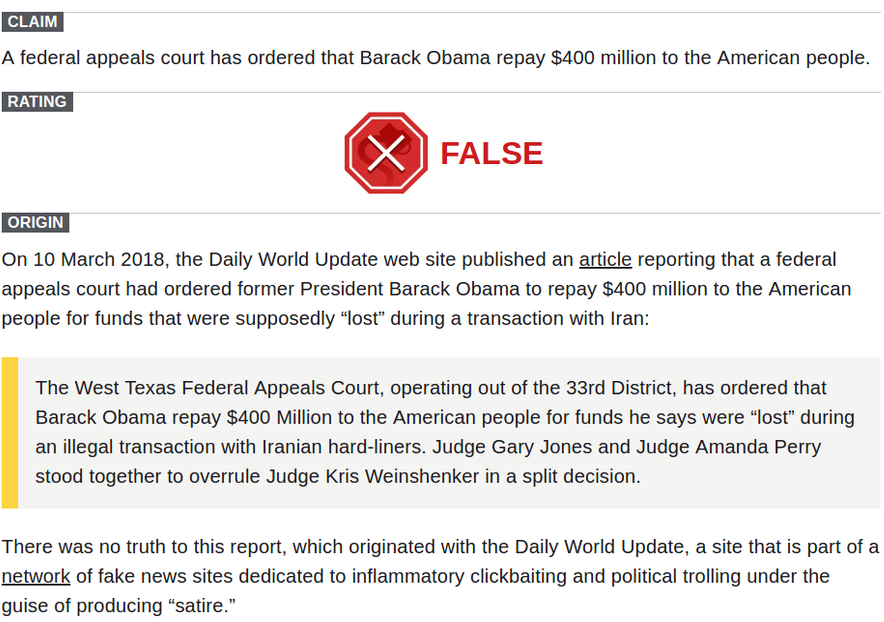}}
\vspace{-3.0ex}
\caption{Snopes fact-checking data example}
\label{fig:snopes}
\vspace{-1.5ex}
\end{figure}

Snopes is a large-scale fact-checking platform that employs human fact-checkers to validate claims. 
A simple \emph{fact-checking instance} from the Snopes website is shown in Figure~\ref{fig:snopes}. 
% As seen in the figure, each instance is represented by a set of fields which store information relevant for fact-checking. 
At the top of the page, the \emph{claim} and the \emph{verdict} (rating) are given. 
The fact-checkers additionally provide a \emph{resolution} (origin), which backs up the verdict. 
Evidence in the resolution, which we call \emph{evidence text snippets} (ETSs), is marked with a yellow bar.
As additional validation support, Snopes fact-checkers provide URLs\footnote{underlined words in the resolution are hyperlinks} for \emph{original documents} (ODCs)
from which the ETSs have been extracted or which provide additional information.

Our crawler extracts the \emph{claims}, \emph{verdicts}, \emph{ETSs}, the \emph{resolution}, as well as \emph{ODCs} along with their URLs, 
thereby enriching the ETSs with useful contextual information.
Snopes is almost entirely focused on claims made on English speaking websites.
Our corpus therefore only features English fact-checking instances.

%%%%%%%%%%%%%%%%%%%%%%%%%%%%%%%%%%%%%%%%%%%%%%%%%%%%%%%%%%%%%%%
\subsection{Corpus annotation}
\label{sec:corp_ann}

While ETSs express a stance towards the claim, which is useful information for the fact-checking process,
this stance is not explicitly stated on the Snopes website. 
Moreover, the ETSs given by fact-checkers are quite coarse and 
often contain detailed background information that is not directly related to the claim and consequently not useful for its validation.
In order to obtain an informative, high-quality collection of evidence, 
we asked crowd-workers to label the stance of ETSs and to extract sentence-level evidence from the ETSs 
that are directly relevant for the validation of the claim. 
We further refer to these sentences as \emph{fine grained evidence} (FGE). 

% task definition and annotation guidelines
\noindent\textbf{Stance annotation.} 
We asked crowd workers on Amazon Mechanical Turk\footnote{\url{https://www.mturk.com/}} to annotate
whether an ETS \emph{agrees} with the claim,
\emph{refutes} it, or has \emph{no stance} towards the claim.
An ETS was only considered to express a stance if it explicitly referred to the claim
and either expressed support for it or refuted it. 
In all other cases, the ETS was considered as having \emph{no stance}.

\noindent\textbf{FGE annotation.} 
We filtered out ETSs with \emph{no stance}, as they do not contain supporting or refuting FGE.
If an ETS was annotated as supporting the claim,
the crowd workers selected only supporting sentences;
if the ETS was annotated as refuting the claim, 
only refuting sentences were selected.
Table~\ref{tbl:stance_example} shows two examples of ETSs with annotated FGE. 
As can be observed, not all information given in the original ETS is directly relevant for validating the claim. 
For example, sentence (1c) in the first example's ETS simply provides additional background information 
and is therefore not considered FGE.

\begin{table}[h]
\begin{center}
\begin{tabular}{ |l| } 
 \hline
 
  \textbf{ETS stance}: support \\ 

  \hline
  
 \textbf{Claim:} The Fox News will be shutting down \\
  for routine maintenance on 21 Jan. 2013.   \\ 
  \hline
  
\textbf{Evidence text snippet:} \\
(1a) \emph{Fox News Channel announced today that} \\
 \emph{ it would shutdown for what it called} \\
  \emph{ ``routine maintenance''.} \\
(1b) \emph{The shutdown is on 21 January 2013.} \\
(1c) Fox News president Roger Ailes explained   \\
 the timing of the shutdown:  ``We wanted  \\
  to pick a time when nothing would be    \\ 
happening that our viewers want to see.'' \\
 \hline
\end{tabular}
\vspace{2mm}
\begin{tabular}{ |l| } 
 \hline
 
  \textbf{ETS stance}: refute \\
  \hline
  
 \textbf{Claim:} Donald Trump supported Emmanuel \\
  Macron during the French election.   \\ 
  \hline
  
\textbf{Evidence text snippet:} \\
(2a) In their first meeting, the U.S. President \\
told Emmanuel Macron that he had been his \\
favorite in the French presidential election \\
saying  ``You were my guy''. \\
(2b) \emph{In an interview with the Associated Press,} \\
\emph{however, Trump said he thinks Le Pen} \\
\emph{is stronger than Macron on what's been going} \\
\emph{on in France.} \\
 \hline
\end{tabular}
\end{center}
\vspace{-3.0ex}
\caption{Examples of FGE annotation in supporting (top) and refuting (bottom) ETSs, sentences selected as FGE in italic.}
\label{tbl:stance_example}
\vspace{-3.0ex}
\end{table}

%%%%%%%%%%%%%%%%%%%%%%%%%%%%%%%%%%%%%%%%%%%%%%%%%%%%%%%%%%%%%%%%%%%%%%%%%%%%%%%%%%%%%%%%%%%%%%%%%%%
\section{Corpus analysis} \label{sec:corpus_analysis}

\subsection{Inter-annotator agreement}
\label{sec:inter_ano}

\noindent\textbf{Stance annotation.} Every ETS was annotated by at least six crowd workers. 
We evaluate the inter-annotator agreement between groups of workers as proposed by \citet{habernal2017argument}, 
i.e. by randomly dividing the workers into two equal groups % (of at least 3 workers)
and determining the aggregate annotation for each group using MACE~\cite{hovy2013learning}. 
The final inter-annotator agreement score is obtained by comparing the aggregate annotation of the two groups.
Using this procedure, we obtain a Cohen's Kappa of $\kappa = 0.7$ \cite{cohen1968weighted},
indicating a substantial agreement between the crowd workers~\citep{artstein2008inter}.
The gold annotations of the ETS stances were computed with MACE, 
using the annotations of all crowd workers. % (at least six for each ETS).
We have further assessed the quality of the annotations performed by crowd workers by comparing them to expert annotations. 
Two experts labeled 200 ETSs, reaching the same agreement as the crowd workers, i.e.~$\kappa = 0.7$. 
The agreement between the experts' annotations and the computed 
gold annotations from the crowd workers is also substantial, $\kappa = 0.683$.

% expert annotation
\noindent\textbf{FGE Annotation.} Similar to the stance annotation, 
we used the approach of \citet{habernal2017argument} to compute the agreement. 
The inter-annotator agreement between the crowd workers in this case is $\kappa = 0.55$ Cohen's Kappa. 
We compared the annotations of FGE in 200 ETSs by experts with the annotations by crowd workers,
reaching an agreement of $\kappa = 0.56$.
This is considered as \emph{moderate} inter-annotator agreement \citep{artstein2008inter}.

In fact, the task is significantly more difficult than stance annotation as sentences may provide only partial evidence for or against the claim. In such cases, it is unclear how large the information overlap between sentence and claim should be for a sentence to be FGE.
The sentence (1a) in Table~\ref{tbl:stance_example}, for example, 
only refers to one part of the claim without mentioning the time of the shutdown.
We can further modify the example in order to make the problem more obvious:
\emph{(a) The channel announced today that it is planing a shutdown. (b) Fox News made an announcement today.}

% \noindent \emph{(a) The Fox News Channel announced today that it is planing a shutdown on 21 January 2013.}
 
% \noindent \emph{c. The maintenance of the Fox News Channel is on 21 January 2013.}
 
% \noindent \emph{(c) Fox News made an announcement today.}

As the example illustrates, there is a gradual transition between sentences 
that can be considered as essential for the validation of the claim
and those which just provide minor negligible details or unrelated information. 
% We therefore considered one or several sentences,
% which cover all important aspects of the claim as legitimate evidence.
% \Cl{}{[The last sentence is unclear. How much information has to be present in a sentence so that the sentence is FGE?]}
Nevertheless, even though the inter-annotator agreement for the annotation of FGE is lower 
than for the annotation of ETS stance, 
compared to other annotation problems \cite{Zechner2002,Benikova2016,tauchmann2018beyond} that are similar to the annotation of FGE, our framework leads to a better agreement.

%%%%%%%%%%%%%%%%%%%%%%%%%%%%%%%%%%%%%%%%%%%%%%%%%%%%%%%%%%%%%
\subsection{Corpus statistics} \label{sec:stat}

Table~\ref{tb:stat} displays the main statistics of the corpus.
In the table, \emph{FGE sets} denotes groups of FGE extracted from the same ETS. 
Many of the ETSs have been annotated as \emph{no stance} (see Table~\ref{tb:etsStnc}) and,
following our annotation study setup, are not used for FGE extraction. 
Therefore, the number of FGE sets is much lower than that of ETSs.
We have found that, on average, an ETS consists of 6.5 sentences.
For those ETSs that have support/refute stance, on average, 2.3 sentences are selected as FGE. 
For many of the ETSs, no original documents (ODCs) have been provided (documents from which they have been extracted).
On the other hand, in many instances, links to ODCs are given that provide additional information, 
but from which no ETSs have been extracted.

% \Cl{}{[Only for those ETS that have support/refute stance? Else, the number 2.3 doesn't make much sense.]}
% \Cl{}{[Would be good to say something about ODC. Is each ODC linked to an ETS? Can an ODC be linked to multiple ETS? Can an ETS have more than one ODC? Right now the number in the table doesn't mean much...]}

\begin{table}[h]
\vspace{-1.0ex}
\centering
\begin{tabular}{l c c c c }
  \toprule
  entity: & claims & ETSs & FGE sets & ODCs     \\ % & ETS with ODC 
  \midrule
  count:    & 6,422 & 16,509  & 8,291 &  14,296    \\ %  & 6,297
  \bottomrule
\end{tabular}
\vspace{-1.5ex}
\caption{Overall statistics of the corpus}
\vspace{-1.5ex}
\label{tb:stat}
\end{table}

The distribution of verdicts in Table~\ref{tb:verdict}
shows that the dataset is unbalanced in favor of false claims. 
The label \emph{other} refers to a collocation of verdicts 
that do not express a tendency towards declaring the claim as being false or true, 
such as \emph{mixture, unproven, outdated, legend,} etc.

\begin{table}[H]
\vspace{-1.0ex}
\centering
\begin{tabular}{l c c c c c}
  \toprule
  verdict: & false & true &  \begin{tabular}{@{}c@{}}most. \\ false\end{tabular}  & \begin{tabular}{@{}c@{}}most. \\ true\end{tabular}  & other  \\
  \midrule
  count    & 2,943 & 659  & 334 & 93 & 2,393  \\ 
  \%    & 45.8 & 10.3  & 5.2 & 1.4 & 37.3  \\ 
  \bottomrule
\end{tabular}
\vspace{-1.7ex}
\caption{Distribution of verdicts for claims}
\vspace{-2.0ex}
\label{tb:verdict}
\end{table}
% \Cl{}{If space needed, delete count or \%, one is enough.}

% ETS FGE analysis -----------------------------------------------------
Table~\ref{tb:etsStnc} shows the stance distribution for ETSs.
Here, supporting ETSs and ETSs that do not express any stance are dominating.

% stance classes distribution
\begin{table}[H]
\vspace{-1.0ex}
\centering
\begin{tabular}{l c c c c }
  \toprule
  stance: & support & refute & no stance   \\
  \midrule
    \textbf{ETSs}: & & & \\
  count & 6,734 & 2,266 & 7,508	  \\ 
  \% & 40.8 & 13.7  & 45.5   \\ 
    \midrule
     \textbf{FGE sets}: & & & \\
    count & 6,178 & 2,113 &	--   \\ 
  \% & 74.5 & 25.5   & --  \\ 
  \bottomrule
\end{tabular}
\vspace{-1.3ex}
\caption{Class distribution of ETSs the FGE sets}
\vspace{-2.0ex}
\label{tb:etsStnc}
\end{table}

For supporting and refuting ETSs 
annotators identified FGE sets for 8,291 out of 8,998 ETSs. 
ETSs with a stance but without FGE sets often miss a clear connection to the claim, so
the annotators did not annotate any sentences in these cases.
The class distribution of the FGE sets in Table~\ref{tb:etsStnc} shows that supporting ETSs are more dominant.

%% Data Statement (biases)
To identify potential biases in our new dataset, we investigated which topics are prevalent by grouping
the fact-checking instances (claims with their resolutions) into categories defined by Snopes.
According to our analysis, the four categories \emph{Fake News}, \emph{Political News}, \emph{Politics} and \emph{Fauxtography}
are dominant in the corpus ranging from more than 700 to about 900 instances.
A significant number of instances are present in the categories \emph{Inboxer Rebellion (Email hoax)}, \emph{Business}, \emph{Medical}, \emph{Entertainment} and \emph{Crime}.

We further investigated the sources of the collected documents (ODCs) 
and grouped them into a number of classes. 
We found that 38\% of the articles are from different news websites ranging from mainstream news like CNN
to tabloid press and partisan news. 
The second largest group of documents are false news and satirical articles with 30\%.
Here, the majority of articles are from the two websites \emph{thelastlineofdefense.org} and \emph{worldnewsdailyreport.com}.
The third class of documents, with a share of 11\%, are from social media like Facebook and Twitter. 
The remaining 21\% of documents come from diverse sources, such as debate blogs, governmental domains, online retail, 
or entertainment websites.

%%%%%%%%%%%%%%%%%%%%%%%%%%%%%%%%%%%%%%%%%%%%%%%%%%%%%%%%%%%%%
\subsection{Discussion} \label{sec:corp_dis}
%% FEVER vs Snopes

I this subsection, we briefly discuss the differences of our corpus to the FEVER dataset
as the most comprehensive dataset introduced so far.
Due to the way the FEVER dataset was constructed, 
the claim validation problem defined by this corpus is different compared to the problem setting defined by our corpus. %  
The verdict of a claim for FEVER depends on the stance of the evidence, that is, 
if the stance of the evidence is agree the claim is necessarily true, 
and if the stance is disagree the claim is necessarily false. 
As a result, the claim validation problem can be reduced to stance detection.
Such a transformation is not possible for our corpus,
as the evidence might originate from unreliable sources and a claim may have both supporting and refuting ETSs. 
The stance of ETSs is therefore not necessarily indicative of the veracity of the claim.
In order to investigate how the stance is related to the verdict of the claim for our dataset,
we computed their correlation. 
In the correlation analysis, we considered how a claims' verdict, represented by the classes \emph{false}, \emph{mostly false}, \emph{other}, \emph{mostly true}, \emph{true}, correlates with the number of supporting ETSs minus the number of refuting ETSs.
More precisely, the verdicts of the claims are considered as one variable, 
which can take 5 discreet values ranging from \emph{false} to \emph{true}, and the stance is considered as the other variable, 
which is represented by the difference between the number of supporting versus the number of refuting evidence.
We found that the verdict is only weakly correlated with the stance, 
as indicated by the Pearson correlation coefficient of 0.16.
This illustrates that the fact-checking problem setting for our corpus is more challenging than for the FEVER dataset.

%%%%%%%%%%%%%%%%%%%%%%%%%%%%%%%%%%%%%%%%%%%%%%%%%%%%%%%%%%
\section{Experiments and error analysis} \label{sec:erran}

%% Clarification
The annotation of the corpus described in the previous section provides supervision
for different fact-checking sub-tasks.
In this paper, we perform experiments for the following sub-tasks: 
(1) detection of the stance of the ETSs with respect to the claim,
(2) identification of FGE in the ETSs, and
(3) prediction of a claim's verdict given FGE.
% (3) validation of the claims on the basis of the verdicts provided by the Snopes fact-checkers 
% and the FGE annotated by crowd workers. \Cl{}{[This sounds as if the verdicts are used for the prediction, but I guess the verdicts are the ground truth?! So maybe better: (3) prediction of a claim's verdict given FGE]}

%% Future work
There are a number of experiments beyond the scope of this paper, 
which are left for future work:
(1) retrieval of the original documents (ODCs) given a claim,
(2) identification of ETSs in ODCs, and
(3) prediction of a claim's verdict on the basis of FGE, the stance of FGE, and their sources.

%% sentence selection
Moreover, in this paper, we consider the three tasks independent of each other rather than as a pipeline. 
In other words, we always take the gold standard from the preceding task instead of the output of the preceding model in the pipeline. 
For the three independent tasks, we use recently suggested models that achieved high performance in similar problem settings.
In addition, we provide the human agreement bound,
which is determined by comparing expert annotations for 200 ETSs 
to the gold standard derived from crowd worker annotations (Section~\ref{sec:inter_ano}).

%-----------------------------------
\subsection{Stance detection}

In the stance detection task, models need to determine whether an ETS \emph{supports} or \emph{refutes} a claim, 
or expresses \emph{no stance} with respect to the claim.

\subsubsection{Models and Results}

%% FNC
We report the performance of the following models: 
\texttt{AtheneMLP} is a feature-based multi-layer perceptron \cite{hanselowski2018retrospective}, 
 which has reached the second rank in the Fake News Challenge. %\footnote{\url{http://www.fakenewschallenge.org/}}
\texttt{DecompAttent} \cite{parikh2016decomposable} 
is a neural network with a relatively small number of parameters that uses decomposable attention,
reaching good results on the Stanford Natural Language Inference task \cite{bowman2015large}.
\texttt{USE+Attent} is a model which uses the Universal Sentence Encoder (USE) \cite{cer2018universal}
to extract representations for the sentences of the ETSs and the claim.
For the classification of the stance, an attention mechanism and a MLP is used.

% performance comparison illustrated
The results in Table~\ref{tb:baseStance} show that \texttt{AtheneMLP} scores highest.
Similar to the outcome of the Fake News Challenge, 
feature-based models outperform neural networks based on word embeddings~\cite{hanselowski2018retrospective}.
As the comparison to the human agreement bound suggests, % \Cl{,}{that} 
there is still substantial room for improvement.

\begin{table}[H]
\vspace{-1.5ex}
\centering
\begin{tabular}{l c c c c c c}
  \toprule
     model  & recall & precision     & F1m    \\
  \midrule
  agreement bound               & 0.770    & 0.837    &    0.802    \\ 
  random baseline               & 0.333    & 0.333    &    0.333    \\ 
  majority vote                 & 0.150    & 0.333    &    0.206    \\ 
  \midrule
  AtheneMLP                     & \textbf{0.585}    & \textbf{0.607}    &    \textbf{0.596}    \\ 
  DecompAttent                  & 0.510    & 0.560    &    0.534    \\ 
    USE+Attent                  & 0.380    & 0.505    &    0.434    \\ 
  \bottomrule
\end{tabular}
\vspace{-1.0ex}
\caption{Stance detection results (F1m = F1 macro)}
\vspace{-2.0ex}
\label{tb:baseStance}
\end{table}

\subsubsection{Error analysis}

% Lexical overlap most significant features
We  performed an error analysis for the best-scoring model \texttt{AtheneMLP}.
The error analysis has shown that \emph{supporting} ETSs are mostly classified correctly 
if there is a significant lexical overlap between the claim and the ETS.
If the claim and the ETSs use different wording,
or if the ETS implies the validity of the claim without explicitly referring to it, 
the model often misclassifies the snippets 
(see example in the Appendix~\ref{sec:app_stan}). % as having \emph{no stance}.
This is not surprising, as the model is based on bag-of-words, topic models, and lexica.

% majority classes support, no stance dominant better classified
Moreover, as the distribution of the classes in Table~\ref{tb:etsStnc} shows, 
\emph{support} and \emph{no stance} are more dominant than the \emph{refute} class.
The model is therefore biased towards these classes and is less likely to predict 
\emph{refute} (see confusion matrix in the Appendix Table~\ref{tb:conf_stan}). 
An analysis of the misclassified \emph{refute} ETSs has shown that the contradiction is often expressed in difficult terms,
which the model could not detect, e.g. 
 ``the myth originated'', ``no effect can be observed'', ``The short answer is no''. % "There is no evidence". %%

%-----------------------------------
\subsection{Evidence extraction}

% As described in Section~\ref{sec:corp_ann}, 
We define evidence extraction 
as the identification of fine-grained evidence (FGE) in the evidence text snippets (ETSs).
The problem can be approached in two ways,
either as a \emph{classification problem}, 
where each sentence from the ETSs is classified as to whether it is an evidence for a given claim,
or as a \emph{ranking problem}, in the way defined in the FEVER shared task.
For FEVER, sentences in introductory sections of Wikipedia articles 
need to be ranked according to their relevance for the validation of the claim
and the 5 highest ranked sentences are taken as evidence.

\subsubsection{Models and Results}

% classification problem setting
We consider the task as a ranking problem, but also provide the human agreement bound, 
the random baseline and the majority vote 
for evidence extraction as a \emph{classification problem} for future reference in Table~\ref{tb:baseFGE} in the Appendix.

% evaluation metric
To evaluate the performance of the models in the \emph{ranking} setup,
we measure the precision and recall on five highest ranked ETS sentences (precision @5 and recall @5),
similar to the evaluation procedure used in the FEVER shared task.
% models
Table~\ref{tb:baseFGErank} summarizes the performance of several models on our corpus. 
The \texttt{rankingESIM}~\cite{hanselowski2018ukp} was the best performing model on the FEVER evidence extraction task. 
The \texttt{Tf-Idf} model \cite{Thorne18Fever} served as a baseline in the FEVER shared task.
We also evaluate the performance of \texttt{DecompAttent} and a simple \texttt{BiLSTM}~\cite{hochreiter1997long} architecture.
% to determine how the models used in the FEVER shared task compare to relatively simple neural networks. 
To adjust the latter two models to the ranking problem setting, 
we used the hinge loss objective function with negative sampling as implemented in the \texttt{rankingESIM} model. 
As in the FEVER shared task, we consider the recall @5 as a metric for the evaluation of the systems.

The results in Table~\ref{tb:baseFGErank} illustrate that, in terms of recall, 
the neural networks with a small number of parameters, \texttt{BiLSTM} and \texttt{DecompAttent}, perform best. 
The \texttt{Tf-Idf} model reaches best results in terms of precision.
The \texttt{rankingESIM} reaches a relatively low score and is not able to beat the random baseline. 
We assume this is because the model has a large number of parameters and requires many training instances. 

\begin{table}[h]
\vspace{-1.0ex}
\centering
\begin{tabular}{l c c }
  \toprule
     model  &  precision @5  & recall @5       \\
  \midrule
      random baseline     & 0.296             & 0.529       \\ 
  \midrule
    BiLSTM              & 0.451             & \textbf{0.637}       \\ 
    DecompAttent        & 0.420             & 0.627       \\ 
    Tf-Idf              & \textbf{0.627}    & 0.601       \\ 
    rankingESIM         & 0.288             & 0.507       \\ 
  \bottomrule
\end{tabular}
\vspace{-1.5ex}
\caption{Evidence extraction: ranking setting}
\vspace{-2.5ex}
\label{tb:baseFGErank}
\end{table}

\subsubsection{Error analysis}

We performed an error analysis for the \texttt{BiLSTM} and the \texttt{Tf-Idf} model, 
as they reach the highest recall and precision, respectively. 
\texttt{Tf-Idf} achieves the best precision 
because it only predicts a small set of sentences, which have lexical overlap with the claim.
The model therefore misses FGE that paraphrase the claim. % or synonyms are used. 
The \texttt{BiLSTM} is better able to capture the semantics of the sentences. 
We believe that it was therefore able to take related word pairs, such as ``\textit{Israel}'' - ``\textit{Jewish}'', ``\textit{price}''-``\textit{sold}'', ``\textit{pointed}''-``\textit{pointing}'',  ``\textit{broken}"-"\textit{injured}'',
into account during the ranking process.
Nevertheless, the model fails when the relationship between the claim and the potential FGE is more elaborate, 
e.g. if the claim is not paraphrased, but reasons for it being true are provided.
An example of a misclassified sentence is given in the Appendix~\ref{sec:ap_sen}.

%-----------------------------------
\subsection{Claim validation}

We formulate the claim validation problem in such a way
that we can compare it to the FEVER \emph{recognizing textual entailment} task.
Thus, as illustrated in Table~\ref{tb:comp}, we compress the different verdicts present on the Snopes webpage 
into three categories of the FEVER shared task. 
In order to form the \emph{not enough information} (NEI) class, 
we compress the three verdicts \emph{mixture}, \emph{unproven}, and \emph{undetermined}.
We entirely omit all the other verdicts like \emph{legend, outdated, miscaptioned},
as these cases are ambiguous and difficult to classify.
For the classification of the claims, we provide only the FGE as they contain the most important information from ETSs.

\begin{table}[h]
\vspace{-1.0ex}
\centering
\begin{tabular}{l l}
  \toprule
     FEVER  &  Snopes        \\
  \midrule
refuted:  &  false, mostly false       \\
supported:  &  true, mostly true      \\
NEI:  &  mixture, unproven, undetermined     \\
  \bottomrule
\end{tabular}
\vspace{-1.5ex}
\caption{Compression of Snopes verdicts }
\vspace{-3.0ex}
\label{tb:comp}
\end{table}

\subsubsection{Experiments}

For the claim validation, we consider models of different complexity: 
\texttt{BertEmb} is an MLP classifier which is based on BERT pre-trained embeddings~\cite{devlin2018bert}; %\footnote{\url{https://github.com/hanxiao/bert-as-service}}; 
\texttt{DecompAttent} was used in the FEVER shared task as baseline; 
%% extendedESIM
\texttt{extendedESIM} is an extended version of the ESIM model ~\cite{hanselowski2018ukp}  
% \Revised{}{is designed to perform textual entailment on the basis of several input sentences and} 
reaching the third rank in the FEVER shared task; 
% \Revised{in the claim classification part of the}{}  
\texttt{BiLSTM} is a simple BiLSTM architecture; 
\texttt{USE+MLP} is the Universal Sentence Encoder combined with a MLP;
\texttt{SVM} is an SVM classifier based on bag-of-words, unigrams, and topic models.

The results illustrated in Table~\ref{tb:baseClaim} show that \texttt{BertEmb}, \texttt{USE+MLP}, \texttt{BiLSTM}, and \texttt{extendedESIM}
reach similar performance, with \texttt{BertEmb} being the best. 
However, compared to the FEVER claim validation problem, where systems reach up to 0.7 F1 macro,
the scores are relatively low.
Thus, there is ample opportunity for improvement by future systems.

\begin{table}%[H]
\centering
\begin{tabular}{l c c c c c c}
  \toprule
     Labeling method   & recall m & prec. m     & F1 m   \\
  \midrule
%   quantBaseline         & 0.415   & 0.377   & 0.395   \\
   random baseline       & 0.333   & 0.333   & 0.333   \\ 
   majority vote         & 0.198   & 0.170   & 0.249   \\ 
 \midrule
   BertEmb          & 0.477   & \textbf{0.493}   & \textbf{0.485}   \\ 
   USE+MLP              & 0.483   & 0.468   & 0.475   \\ 
    BiLSTM               & 0.456   & 0.473   & 0.464   \\ 
   extendedESIM         & \textbf{0.561}   & 0.503   & 0.454 \\ 
   featureSVM           & 0.384   & 0.396   & 0.390   \\ 
   DecompAttent         & 0.336   & 0.312   & 0.324   \\ 
  \bottomrule
\end{tabular}
\vspace{-1.0ex}
\caption{Claim validation results (m = macro)}
\vspace{-3.5ex}
\label{tb:baseClaim}
\end{table}

\subsubsection{Error analysis} \label{sec:claimer}

% negation included though the claim is true 
%% ERORR analysis
We performed an error analysis for the best-scoring model \texttt{BertEmb}. 
The class distribution for claim validation is highly biased towards \emph{refuted} (false) claims
and, therefore, claims are frequently labeled as \emph{refuted} even though they belong to one of the other two classes (see confusion matrix in the Appendix in Table~\ref{tb:conf_claim}).

% which was probably the cause for the misclassification.
We have also found that it is often difficult to classify the claims as the provided FGE in many cases are contradicting 
(e.g. Appendix~\ref{sec:ap_claim}).
Although the corpus is biased towards false claims (Table~\ref{tb:etsStnc}), 
there is a large number of ETSs that support those false claims (Table~\ref{tb:verdict}). 
As discussed in Section~\ref{sec:stat}, 
this is because many of the retrieved ETSs originate from false news websites.

Another possible reason for the lower performance % scores on our corpus 
is that our data is heterogeneous and, therefore, 
it is more challenging for a machine learning model to generalize.
In fact, we have performed additional experiments in which we pre-trained a model on the FEVER corpus 
and fine-tuned the parameters on our corpus and vice versa. 
However, no significant performance gain could be observed in both experiments

Based on our analysis, we conclude that 
heterogeneous data and FGE from unreliable sources, as found in our corpus and in the real world,
make it difficult to correctly classify the claims. 
Thus, in future experiments, not just FGE need to be taken into account, but also additional information from our newly constructed corpus,
that is, the stance of the FGE, FGE sources, 
and documents from the Snopes website which provide additional information about the claim.
Taking all this information into account would enable the system to find a consistent configuration of these labels  
and thus potentially help to improve performance.
For instance, a claim that is \emph{supported} by evidence coming from an \emph{unreliable} source is most likely \emph{false}.
In fact, we believe that modeling the meta-information about the evidence 
and the claim more explicitly represents an important step in making progress in automated fact-checking.

%%%%%%%%%%%%%%%%%%%%%%%%%%%%%%%%%%%%%%%%%%%%%%%%%%%%%%%%%%
\section{Conclusion} \label{sec:conc}
In this paper, we have introduced a new richly annotated corpus 
for training machine learning models for the core tasks in the fact-checking process. 
The corpus is based on heterogeneous web sources, such as blogs, social media, and news, where most false claims originate. 
It includes validated claims along with related documents, evidence of two granularity levels, 
the sources of the evidence, and the stance of the evidence towards the claim.
This allows training machine learning systems for document retrieval, stance detection, evidence extraction, and claim validation.

We have described the structure and statistics of the corpus, 
as well as our methodology for the annotation of evidence and the stance of the evidence. 
We have also presented experiments for stance detection, evidence extraction, and claim validation
with models that achieve high performance in similar problem settings. 
%%% error analysis
In order to support the development of machine learning approaches that go beyond the presented models,
we provided an error analysis for each of the three tasks, identifying difficulties with each.

Our analysis has shown that the fact-checking problem defined by our corpus is more difficult than for other datasets.
Heterogeneous data and evidence from unreliable sources, as found in our corpus and in the real world, 
make it difficult to correctly classify the claims. 
We conclude that more elaborate approaches are required to achieve higher performance in this challenging setting.

%%%%%%%%%%%%%%%%%%%%%%%%%%%%%%%%%%%%%%%%%%%%%%%%%%%%%%%%%%%%%%%%%%%%%%%%%%%%%
\section{Acknowledgements}

This work has been supported by the German Research Foundation as part of the Research
Training Group ''Adaptive Preparation of Information from Heterogeneous Sources'' (AIPHES) at the Technische Universit\"at Darmstadt under grant No. GRK 1994/1.

%%%%%%%%%%%%%%%%%%%%%%%%%%%%%%%%%%%%%%%%%%%%%%%%%%%%%%%%%%%%%%%%%%%%%%%%%%%%%%%%%%%

\bibliography{emnlp-ijcnlp-2019}
\bibliographystyle{acl_natbib}

\vspace{11cm}

%%%%%%%%%%%%%%%%%%%%%%%%%%%%%%%%%%%%%%%%%%%%%%%%%%%%%%%%%%%%%%%%%%%%%%%%%%%%%
\appendix

\pagebreak
.
\pagebreak

\section{Appendix}

\subsection{Evidence extraction classification problem: baselines and agreement bound}

\begin{table}[h]
\centering
\begin{tabular}{l r r r}
  \toprule
     model  & recall m & precis. m & F1 m    \\
  \midrule
    agreement bound    & 0.769     & 0.725   &  0.746   \\ 
    random baseline     & 0.500       & 0.500      & 0.500      \\ 
     majority vote      & 0.343     & 0.500    & 0.407    \\ 
  \bottomrule
\end{tabular}
\vspace{-1.5ex}
\caption{Evidence extraction classification problem: baselines and agreement bound (m = macro)}
\vspace{-1.5ex}
\label{tb:baseFGE}
\end{table}

\subsection{Error analysis}

\subsubsection{Stance detection}
\label{sec:app_stan}

Below we give an instance of a misclassified ETS.
Even though the ETS supports the claim, the lexical overlap is relatively low.
Most likely, for this reason, the model predicts \emph{refute}.

% example 
\noindent Example: 
\noindent\rule{6cm}{0.4pt}
\noindent  Claim: \emph{The Reuters news agency has proscribed the use of the word 'terrorists' to describe those who pulled off the September 11 terrorist attacks on America.}\\
\noindent ETS: 
\emph{Reuters' approach doesn't sit well with some journalists,  who say it amounts to self-censorship. 
``Journalism should be about telling the truth. And when you don't call this a terrorist attack, you're not telling the truth,'' 
 says Rich Noyes, director of media analysis at the conservative Media Research Center. ...
 }\noindent\rule{7.6cm}{0.4pt}
% They also argue that it's inaccurate. 
% ``A news organization's responsibility is to find the facts ... not to play politics with its reporting.''

\begin{table}[H]
\centering
\begin{tabular}{l  c  c  c }
  \toprule
    model $\backslash$ gold  & support & refute & no stance    \\
  \midrule
  support                 & \textbf{472}   &   \textbf{86}   &   175   \\ %\hline  
  refute                  & 41             &  80             &     51   \\ %\hline  
  no stance               &   141          &  74             &    \textbf{531}    \\
  \bottomrule
\end{tabular}
\vspace{-1.5ex}
\caption{Stance detection confusion matrix (\texttt{AtheneMLP}) }
\vspace{-1.5ex}
\label{tb:conf_stan}
\end{table}

\subsubsection{Evidence extraction} \label{sec:ap_sen}

The model wrongly predicts sentences when the topic of the sentences is similar to the topic of the claim, 
but the sentence is not relevant for the validation of the claim:

\noindent Example: 
\noindent\rule{6cm}{0.4pt}
\noindent Claim: \emph{The Department of Homeland Security uncovered a terrorist plot to attack Black Friday shoppers in several locations.}\\
\noindent FGE: \emph{Bhakkar Fatwa is a small, relatively unknown group of Islamic militants and fanatics that originated in Bhakkar Pakistan as the central leadership of Al Qaeda disintegrated under the pressures of U.S. military operations in Afghanistan and drone strikes conducted around the world.}\rule{6cm}{0.4pt}

\subsubsection{Claim validation}
\label{sec:ap_claim}

The FGE are contradicting and the classifier predicts \emph{refuted} instead of \emph{supported}.

\noindent Example: 
\noindent\rule{6cm}{0.4pt}
\noindent Gold standard: \emph{supported}; Prediction: \emph{refuted} 
\noindent  Claim: \emph{As a teenager, U.S. Secretary of State Colin Powell learned to speak Yiddish while working in a Jewish-owned baby equipment store.}\\
\noindent FGE: \emph{As a boy whose friends and employers at the furniture store were Jewish, Powell picked up a smattering of Yiddish.
He kept working at Sickser's through his teens, ... picking up a smattering of Yiddish ... % earning 75 cents an hour and
A spokesman for Mr. Powell said he hadn't heard about the spoof ... } %but confirmed that Gen. Powell does speak a little Yiddish.
\noindent\rule{7.6cm}{0.4pt}
% The low performance of the model was reflected in the examples of the error analysis as it was often difficult to identify,
% why the classifier miss classified a particular instance.

\begin{table}[H]
\centering
\begin{tabular}{l  c  c  c }
  \toprule
    model $\backslash$ gold  & supported & refuted     & NEI   \\
  \midrule
  supported                    & 36   &   26   &    13  \\   
  refuted                      &  \textbf{38}  & \textbf{203}     &  \textbf{53}       \\   
  NEI                         & 18   &   42    &    27    \\
  \bottomrule
\end{tabular}
\vspace{-1.0ex}
\caption{Confusion matrix for claim validation \texttt{BertEmb} (NEI: not enough information)}
\vspace{-2.5ex}
\label{tb:conf_claim}
\end{table}

\end{document}